\documentclass[journal]{IEEEtran}

\usepackage[backend = biber, style = ieee]{biblatex}
\usepackage{authblk}
\usepackage{courier}
\usepackage{graphicx}
\usepackage{fancyhdr}
\usepackage{import}
\usepackage{subfig}
\usepackage{amsmath}
\usepackage{hyperref}
\usepackage[ruled, vlined]{algorithm2e}
\usepackage{tikz}
\usepackage{float}
\usepackage[justification=centering]{caption}
\usepackage{xcolor}
\usepackage[normalem]{ulem}
\usepackage{dblfloatfix}
\usepackage{dirtytalk}

\useunder{\uline}{\ul}{}
\usetikzlibrary{external}
\usetikzlibrary{automata, positioning, arrows}
\title{
    \huge ARGOS: An Automaton Referencing Guided Overtake System \\
    for Head-to-Head Autonomous Racing }

\author[1]{Varundev Sukhil}
\author[2]{Madhur Behl}
\affil[1]{varundev@virginia.edu, PhD Candidate in Computer Engineering, University of Virginia}
\affil[2]{madhur.behl@virginia.edu, Associate Professor of Computer Science, University of Virginia}

\begin{document}

\maketitle

\begin{abstract}
Autonomous overtaking at high speeds is a challenging multi-agent robotics research problem.
The high-speed and close proximity situations that arise in multi-agent autonomous racing require designing algorithms that trade off aggressive overtaking maneuvers and minimize the risk of collision with the opponent.
In this paper, we study a special case of multi-agent autonomous race, called the head-to-head autonomous race, that requires two racecars with similar performance envelopes.
We present a mathematical formulation of an overtake and position defense in this head-to-head autonomous racing scenario, and we introduce the Automaton Referencing Guided Overtake System (ARGOS) framework that supervises the execution of an overtake or position defense maneuver depending on the current role of the racecar.
The ARGOS framework works by decomposing complex overtake and position-defense maneuvers into sequential and temporal submaneuvers that are individually managed and supervised by a network of automatons.
We verify the properties of the ARGOS framework using model-checking and demonstrate results from multiple simulations, which show that the framework meets the desired specifications.
The ARGOS framework performs similar to what can be observed from real-world human-driven motor sport racing.
\end{abstract}

\section{Introduction}
\label{sec:introduction}
High-speed autonomous racing can be considered as a grand challenge for multi-agent robotics and for autonomous vehicles.
Therefore, making progress in this area has the potential to enable breakthroughs in agile and safe autonomy.
To succeed at autonomous racing, an autonomous vehicle is required to perform both precise steering and throttle maneuvers in a physically complex, uncertain environment by executing a series of high-frequency decisions.
At the time of this paper, autonomous racing is becoming a motorsport featuring head-to-head battles of algorithms. 
Roborace~\cite{roborace} claims to feature fully autonomous race cars in the near future, and autonomous racing competitions, such as F1/10 racing~\cite{f1tenth_main, f1tenth_sim}, Formula SAE Driverless, and Indy Autonomous Challenge~\cite{IAC_1}, are both figuratively and literally getting a lot of traction and becoming proving grounds for testing perception, planing, and control algorithms at high speeds.

However, research in autonomous racing has largely focused on a single-agent time-trial style of racing in which a single autonomous racecar completes a lap in the shortest time. 
Time trial poses a number of challenges in terms of dynamic modeling, on-board perception, localization and mapping, trajectory generation and optimal control, but these challenges have largely been addressed at events like the Indy Autonomous Challenge~\cite{IAC_1, IAC_2}
Limited attention has been paid to multi-agent autonomous racing.
Multi-agent autonomous racing is especially difficult since in addition to challenges in dynamic modeling of the vehicles at the limits of control and fast trajectory generation, it also requires state-estimation for other agents, and maneuvers for opportunistic passing while avoiding collisions, along with other objectives such as lap time, boost energy, etc.  
In general, multi-agent autonomous racing provides the opportunity for developing and testing more widely applicable non-cooperative multirobot strategies.
In this paper, we focus on a special case of multi-agent autonomous racing: the head-to-head autonomous race.
Multi-agent racing can be construed as parallel head-to-head races where a racecar is operating under a head-to-head race with one racecar where the ego-racecar is trying to overtake, while simultaneously the ego-racecar is engaged in defending its race rank in another head-to-head race with a different racecar.
Solving the problem of head-to-head autonomous racing is the necessary stepping stone toward a true multi-agent autonomous race.
We make the following contributions in this paper:
\begin{itemize}
    \item We present a modular autonomous head-to-head racing framework with specific guidelines on integrating/adapting components within the framework.
    \item We introduce the ARGOS automaton network - a set of three interconnected automatons that perform overtaking and position defense maneuvers.
    \item We present a model-checking approach to verify that the ARGOS framework is capable of meeting the specifications described in the race rules.
\end{itemize}

The paper is organized in the following manner: (a) we introduce the relevant work in autonomous racing (see Section~\ref{sec:related_work}), (b) we formulate a mathematical model of a head-to-head autonomous racing scenario (see Section~\ref{sec:h2h_problem}), (c) we describe our solution in the form of an autonomous racing framework (see Section~\ref{sec:ARGOS_framework}), (d) we briefly describe how we used formal methods to verify our proposed framework (see Section~\ref{sec:model_checking}), and (e) we present our findings through experiments (see Section~\ref{sec:experiments}).

\section{Related Work in Autonomous Racing}
\label{sec:related_work}
An autonomous racecar's race-specific software stack has three broad categories: (a) perception, (b) planning, and (c) control.
In this section we describe our implementation of the autonomous racing software stack, and the various related and tangential works for each.

\subsection{Perception}

The perception stack for an autonomous racecar helps the racecar understand the environment, the states of itself, and the other racecars.
Generally, perception in autonomous racing includes Simultaneous Localization and Mapping (SLAM)~\cite{slam_1, slam_2} where multiple sensors including Camera, LiDAR, GPS and RADARS work together using techniques such as Normalized Distance Transformation (NDT)~\cite{ndt_mapping} to estimate its own odometry (pose, rotation and velocity), and Convolutional Neural Network (CNN) based image segmentation~\cite{nvidia_cnn} to estimate pose of an opponent racecar.
A simulator can provide ground truth information about the racecar's states, and since the major focus of this paper is on decision making, path planning, and controls, we decided to use the ground truth information from simulation.
In this paper, we extensively used the LGSVL simulator~\cite{lgsvl} with a realistic model of an AV-21 autonomous racecar.
Some other simulators that we considered included CARLA~\cite{carla}, TORCS~\cite{torcs}, and AirSim~\cite{airsim}.
We chose the LGSVL simulator because it included a realistic model of the AV-21 autonomous racecar and a virtual replica of Indianapolis Motor Speedway.

\subsection{Path Planning}

A path is defined as a continuous set of waypoints with a corresponding velocity set point, that is, $W = \{x_i, y_i, v_i\} \forall i \in [1, N]$, where $\{x_i, y_i\}$ and $v_i$ are the pose and velocity set points, respectively, for the waypoint $w_i$ in a path $W$ of length $N$.
Path planning in autonomous racing involves finding the optimal trajectory that satisfies certain race objectives.
Generally, these objectives are divided into two categories: (a) the global objectives, which are handled by the global planner, and (b) the local objectives, which are handled by the local planner.
The global objectives of an autonomous racecar is to find the optimal global raceline with the least lap time~\cite{time_optimal_planning}, minimum curvature throughout the track~\cite{min_curvature_planning}, and the fastest average lap speed using the friction map method~\cite{friction_map_method}. 
Global path planning has been studied exhaustively; therefore, this paper focuses on local path planning and adapts the work done by the authors cited in this section.

Local path planning for autonomous racing is described in Section~\ref{sec:h2h_problem}, and generally involves creating a set of guide control points that satisfy a defined objective (in our case: overtaking or position defense).
In this paper, we use the quintic spline fitting method to construct a spline using a set of guide control points.
Equation~\ref{eq:quintic_spline} shows the general description of a quintic polynomial, and a quintic spline is stitched together using this polynomial.

\begin{equation}
    s(t) = a_0 + a_1t + a_2t^2 + a_3t^3 + a_4t^4 + a_5t^5
\label{eq:quintic_spline}
\end{equation}

Parameters $a_0-a_5$ of the polynomial determine the characteristics of the spline.
Using the known initial states of the racecar including position ($x_s$), velocity ($v_s$), and current acceleration ($a_s$), we can derive the first three parameters at zero time, as shown in Equation~\ref{eq:quintic_params_1}.

\begin{equation}
    \begin{split}
        s(0) & = a_0 = x_s \\
        \dot{s(0)} & = a_1 = v_s \\
        \ddot{s(0)} & = 2a_2 = a_s
    \end{split}
\label{eq:quintic_params_1}
\end{equation}

Path planners parameterize the end time of a local maneuver ($T$), along with the desired final state of the racecar at this end time of the maneuver, including position ($x_e$), velocity ($v_e$), and final acceleration ($a_e$).
Using this information, we can compute the remaining three unknown parameters of the quintic polynomial, as shown in Equation~\ref{eq:quintic_params_2}.

\begin{equation}
    \begin{bmatrix}
        T^3 & T^4 & T^5 \\
        3T^2 & 4T^3 & 5T^4 \\
        6T & 12T^2 & 20T^3
    \end{bmatrix}
    \begin{bmatrix}
        a_3 \\
        a_4 \\
        a_5
    \end{bmatrix} = 
    \begin{bmatrix}
        x_e - x_s - v_sT - \frac{a_sT^2}{2} \\
        v_e - v_s - a_sT \\
        a_e - a_s
    \end{bmatrix}
\label{eq:quintic_params_2}
\end{equation}

Equations~\ref{eq:overtake_spline_control_points} and~\ref{eq:position_defense_spline_control_points} describe how control points of the overtake and position defense splines, respectively, are generated.

\subsection{Path Tracking (Control)}

The control components of an autonomous racing stack use the path generated by the path planners and command the racecar to follow the path.
There are various techniques used for control, and these techniques are broadly classified as (a) mode-free techniques (eg: pure-pursuit control~\cite{pure_pursuit_1, pure_pursuit_2}, Stanley control~\cite{stanley_control}, and rear wheel feedback control~\cite{rear_wheel_feedback} etc.) and (b) model-based control (model predictive control~\cite{mpc_1, mpc_2}, linear quadratic control~\cite{lqr_car}, and Q-learning controller~\cite{Q_control} etc.).

\begin{figure}[h!]
    \centering
    \includegraphics[width=0.65\linewidth]{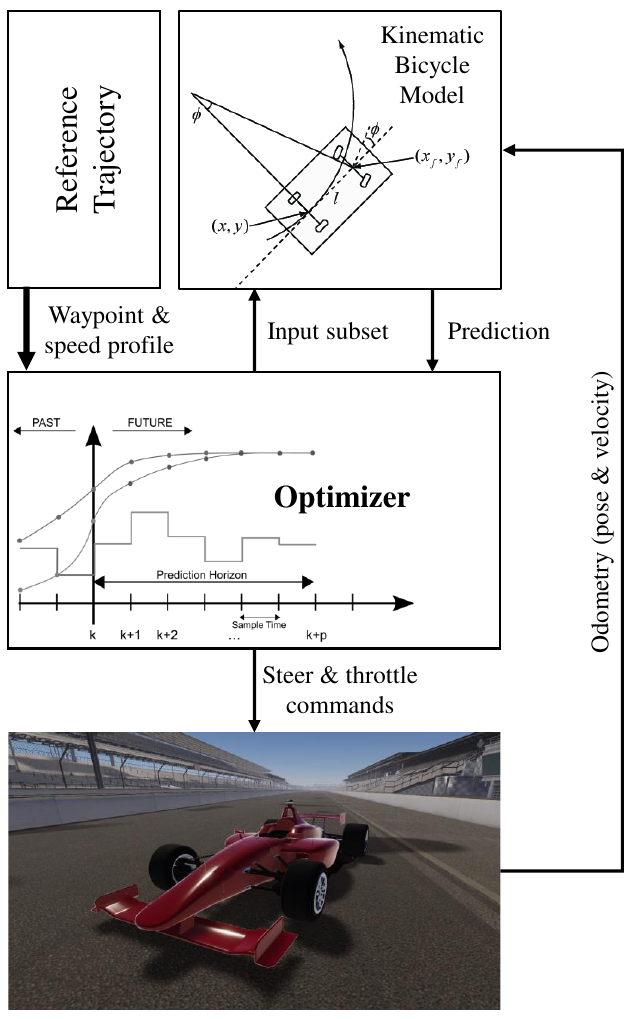}
    \caption{A block diagram view of the MPC based path tracker. Using the kinematic bicycle model of the Dallara AV-21 racecar, the path tracker can accurately determine the optimal steering and throttle to track a reference trajectory.}
    \label{fig:kin_MPC_diagram}
\end{figure}

In this paper, we use a model predictive controller that computes steering, throttle, and brake commands using predictions made by a kinematic bicycle mode of the autonomous racecar to track the reference trajectory.
The linearized model of the racecar used by the path tracker is shown in Equation~\ref{eq:racecar_dynamics}.
The input to the controller is the acceleration $a$ subjected to the acceleration limits $a \epsilon [a_{min}, a_{max}]$, and the steering angle $\delta$ subjected to the steering limits $\delta \epsilon [\delta_{min}, \delta_{max}]$.
In Equation~\ref{eq:racecar_dynamics}, $\{x, y\}$ is the position of the racecar in the global frame, $v$ is the current velocity of the racecar and $\phi$ is the current global heading of the racecar.

\begin{equation}
    \begin{split}
        \dot{x} & = x \cos{\phi} \\
        \dot{y} & = y \sin{\phi} \\
        \dot{v} & = a \\
        \dot{\phi} & = v \frac{\tan{\delta}}{L} 
    \end{split}
\label{eq:racecar_dynamics}
\end{equation}

Figure~\ref{fig:kin_MPC_diagram} shows the block diagram of the kinematic bicycle model used in~\ref{eq:racecar_dynamics}.
The MPC optimizer chooses inputs that minimize the lateral (distance from the race track) and longitudinal (difference from the set velocity profile) components from all predictions.
The controller function is shown in Equation~\ref{eq:mpc_optimizer} where the maneuver cost \textbf{C} is minimized, and \textbf{Q}, \textbf{u}, \textbf{R}, \textbf{z} are the state cost, input, input cost, and the discretized state of the racecar respectively.
Also in Equation~\ref{eq:mpc_optimizer} \textbf{T} is the prediction horizon, \textbf{t} is the current prediction and \textbf{ref} is the tracked reference trajectory and speed profile.
Thus, $z_{T,ref}$ and $z_{t,ref}$ are the controller's reference and prediction at $t$, respectively.

\begin{equation}
    \begin{split}
        C_{min} & = Q_f(z_{T,ref}-z_T)^2 + Q\sum(z_{t,ref}-z_t)^2 \\
                & + R\sum u_t^2 + R_d\sum(u_{t+1}-u_t)^2
    \end{split}
\label{eq:mpc_optimizer}
\end{equation}

\subsection{Advanced Energy Management System}

The Advanced Energy Management System (AEMS), as described in~\cite{autopass_gen_1}, provides a temporary ability for a racecar to exceed its designed top speed during an overtake or close the gap with the opponent.
The limited time for which the racecar can be "boosted" is called the boost-energy budget.
In this paper, each racecar is allowed to use the boost-energy budget only in the passing zones (the front and the back stretch of the virtual race-track), and for a total time of around 20 seconds each lap, at a drain-rate chosen by the racecar.

\section{Head-to-Head Racing: Problem Formulation}
\label{sec:h2h_problem}
\begin{figure*}[h!]
    \centering
    \includegraphics[width=\linewidth]{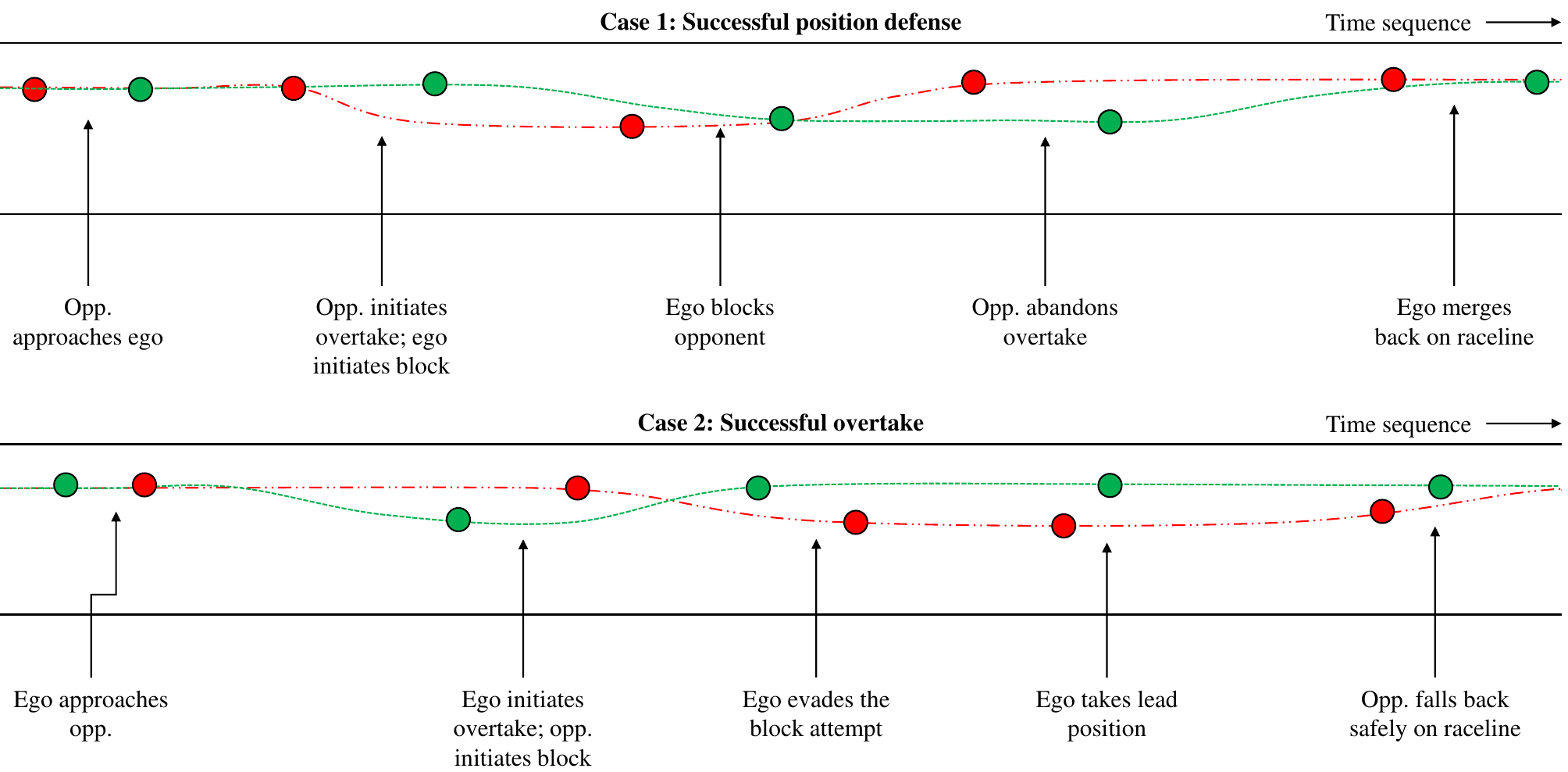}
    \caption{A timed odometry trace for: [Top] a successful position defense, and [Bottom] a successful overtake attempt. In both cases, the red and green dots represent the opponent and ego racecar respectively. Each set of ego and opponent poses represent the sub-maneuvers involved in an overtake attempt or position defense.}
    \label{fig:maneuver_trace_diagram}
\end{figure*}

\begin{figure*}[b!]
    \centering
    \includegraphics[width=0.8\linewidth]{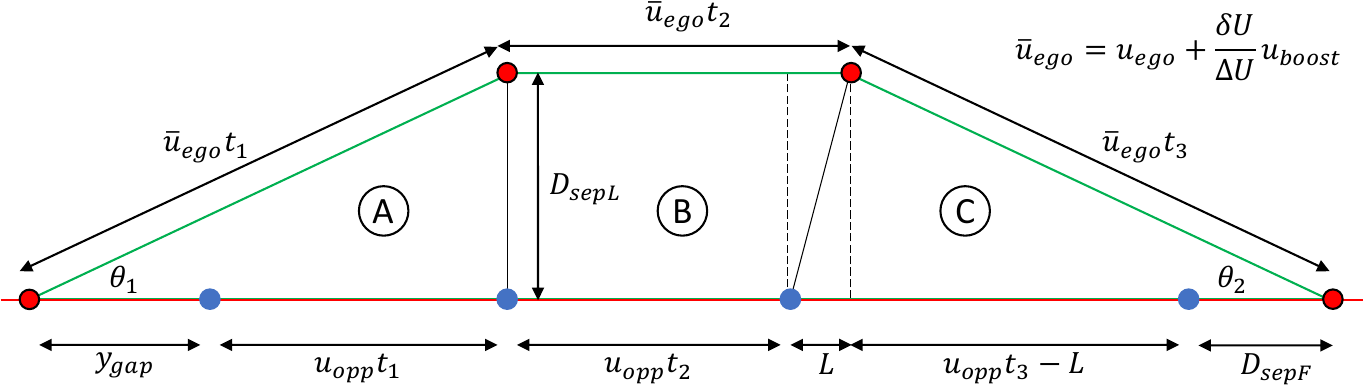}
    \caption{The piecewise trapezoidal overtake geometry model showing the three submaneuvers involved in an overtake [left-right] (A) initiate overtake, (B) pass opponent's car-length, and (C) safely merge in-front of the opponent.}
    \label{fig:overtake_geometry}
\end{figure*}

As discussed in Section~\ref{sec:introduction}, Head-to-Head Autonomous Racing (H2H) is a subset of [Multi-Agent] Autonomous Racing.
Therefore, we model an H2H problem with the assumption that the two racecars involved in the H2H scenario are not affected by other racecars.
In this scenario, each racecar occupies one role: a \textbf{Attacker} and a \textbf{Defender}.
Similarly to the namesake, the defender racecear is the racecar that currently leads the race and is expected to "defend" its leader position, while the attacker is the racecar that is expected to "overtake" the race leader.

The roles of Attacker and Defender are modeled differently, since their expected outcomes are different.
In each role, the problem consists of an "ego" and an "opponent" racecar.
The ego racecar has its "race stack" (defined as a set of nodes that perform perception, path planning, and control tasks, and explained in detail in Section~\ref{sec:related_work}) open to modification by an external method, while the opponent racecar can only be observed in relation to the ego racecar.

\subsection{Requirements \& Expected Behavior}

The \textbf{Requirements} are a set of governing rules (a.k.a. race rules) imposed on the modeling of the attacker and the Defender.
In this paper, we define the following requirements, inspired by~\cite{IAC_1}, and outline the expected behavior of the racecars for each role.

\label{rule:observation_radius}
\noindent \textbf{Rule R1 - Observation Radius (meters):} The maximum distance to which the ego-racecar can track the opponent racecar.
This distance is intentionally set higher for the attacker compared to the Defender, so we expect the Defender to react quickly to an overtake attempt.
Once the racecars are beyond the Attacker's radius, any current maneuver is considered complete (and can be a successful or failed attempt).

\label{rule:block_attempts}
\noindent \textbf{Rule R2 - Blocking Attempt (N):} The maximum number of times the Defender can attempt to block during any overtake.
If the Defender was not limited in blocking attempts, the racecar that started the race in the leader position would remain in that position.
The attacker is expected to be robust in circumventing a block attempt, or lose momentum and disengage.

\label{rule:safety_distance}
\noindent \textbf{Rule R3 - Safety Distance (meters):} The minimum distance between the two closest points in the extrinsic footprint of the racecar.
Racecar controllers have a margin of error when tracking a raceline at high speeds. Safety distance ensures that close proximity maneuvers do not result in a collision.
The attacker is responsible for ensuring that this rule is not violated.
We expect the attacker to have large overtake trajectories around the Defender. We also expect the Defender to execute close-proximity maneuvers and force the attacker to abandon an overtake.

\label{rule:boost_energy}
\noindent \textbf{Rule R4 - Boost Energy (seconds):} Each racecar is given a fixed energy budget that enables it to travel at higher than maximum speed for a limited time.
This boost energy increases the possibility of an overtake and is allowed only on dedicated sections of the track where an overtake can be difficult (e.g., straight sections).
We expect an attacker to extensively use boost energy to perform an overtake.

\label{rule:maneuver_fatigue}
\noindent \textbf{Rule R5 - Maneuver Fatigue (meters):} The maximum distance a racecar can travel in an overtake attempt or block attempt.
Race cars may remain in a high-speed state and continuously drain their allocated boost energy in the first overtake attempt.
To discourage wasteful resource utilization, racecars are allowed a fixed linear distance to complete a maneuver. An attacker is allowed significantly higher overtake distance due to the nature of an overtake attempt.
We expect the Defender to execute a late block and the Attacker to abandon an overtake if the Defender chooses to use boost energy to maintain its position.

\subsection{Modelling Head-to-Head Problem Autonomous Racing}

In this paper, we model the Head-to-Head (H2H) autonomous racing problem with the help of the kinematic bicycle model of the racecar and a cubic spline planner described in Section~\ref{sec:related_work}.
The problem model is subject to the rules described in \textbf{R1 - R5}.
The model is further divided into independent solutions to address overtaking and position defense.
The general parameters for both models include a global race line ($W$) and the race track limits ($B$ including the left limits $B_L$ and the right limits $B_R$) described in Equation~\ref{eq:raceline_and_track_bounds}, where $N$ is the traversal distance of the race track within the race track, and $L$ is the distance from the race track.

\begin{figure*}[h!]
    \centering
    \includegraphics[width=\linewidth]{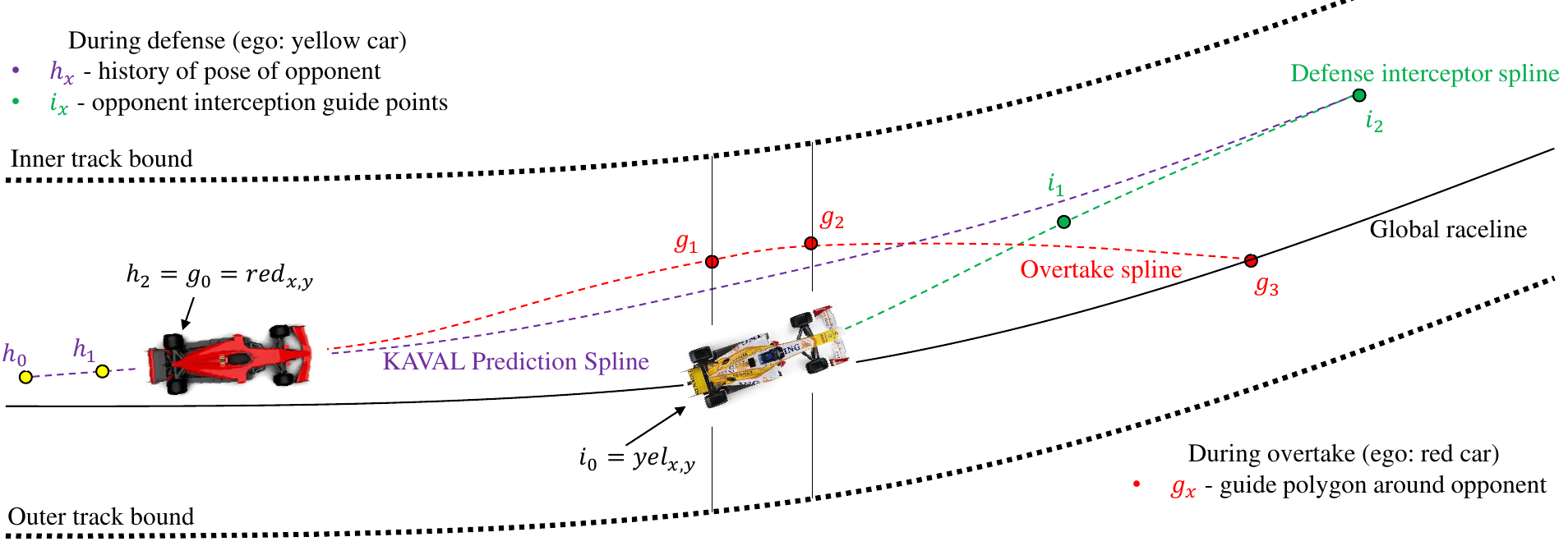}
    \caption{An overview of the quintic spline generation process used in this paper to produce a local overtake (red) and position defense splines (green) using the corresponding set of guide control points.}
    \label{fig:quintic_planner_details}
\end{figure*}

\begin{figure*}[b!]
    \centering
    \includegraphics[width=0.8\linewidth]{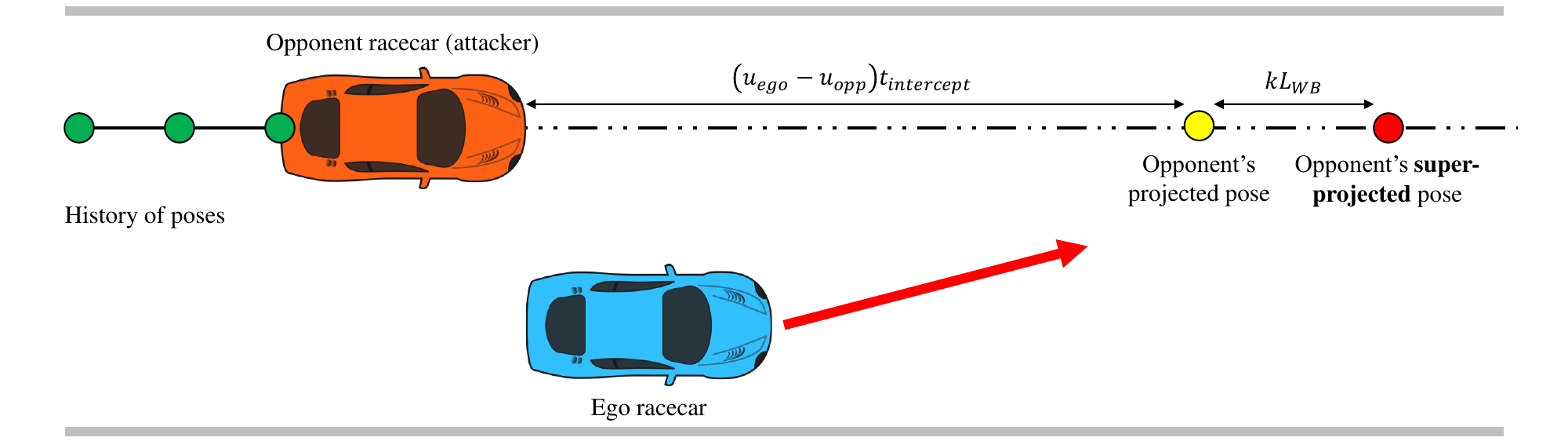}
    \caption{Geometric view of superprojection intercepting position-defense model. The ego racecar must find a pose in the opponent's superprojection and occupy that pose before the opponent to successfully block the opponent.}
    \label{fig:interceptor_geometry}
\end{figure*}

\begin{equation}
    \begin{split}
        W & = \{w_i \langle x_i, y_i, v_i \rangle \forall i\epsilon[0, N]\} \\
        B \langle B_L, B_R \rangle & = \{b_i \langle x_i, y_i \rangle \forall i\epsilon[0, L]\}
    \end{split}
    \label{eq:raceline_and_track_bounds}
\end{equation}

Equation~\ref{eq:helper_funtions} provides an overview of helper functions that provide a tuple $\langle x, y \rangle$ in the relative coordinate system within the known race track boundaries.
The function $g(o_1, o_2)$ computes the point on the spline $o_1$ that is closest to the spline $o_2$.
The function $h(o_1, o_2, d)$ calculates the distance $d$ from the point $o_1$ on the shortest line joining $o_1$ and $o_2$.
Function $j(o_1, o_2, o_3)$ determines if $o_1$ is closer to $o_2$ or $o_3$.
Finally, the function $k(o_1, \theta, d)$ calculates the distance $d$ from the point $o_1$ along the relative angle $\theta$.

\begin{equation}
    \begin{split}
        g(o_1, o_2) & = \min|o_2 - o_1|\langle x, y \rangle \\
        h(o_1, o_2, d) & = \langle o_1(x) + d\cos\theta, o_1(y) + d\sin\theta \rangle \\
        & \Rightarrow \theta = \tan^{-1}(o_2/o_1) \\
        j(o_1, o_2, o_3) & = 
            \begin{cases}
                o_2, & |o_1-o_2| > |o_1-o_3\\
                o_3, & otherwise \\
            \end{cases} \\
        k(o_1, \theta, d) & = \langle o_1(x) + d\cos\theta, o_1(y) + d\sin\theta \rangle
    \end{split}
    \label{eq:helper_funtions}
\end{equation}

\subsubsection{Modelling an Overtake}

The piecewise trapezoidal overtake geometry model described in Figure~\ref{fig:overtake_geometry} shows the sections into which an overtake is broken down in this paper.
Each section represents a submaneuver, and each submaneuver must be executed in the A-B-C sequence as shown in this figure.
The sum of time ($t_i$) associated with each submaneuver must be less than the total boost energy available to the attacker for a successful overtake.
Equation~\ref{eq:total_overtake_path} describes the submaneuver times.

\begin{equation}
    \begin{split}
    t_{A} & = \frac{y_{gap}}{(u_{ego} - u_{opp} + \frac{\delta U}{\Delta U}u_{boost}) cos\theta_{1}} \\
    t_{B} & = \frac{L}{(u_{ego} - u_{opp} + \frac{\delta U}{\Delta U}u_{boost})} \\
    t_{C} & = \frac{D_{sepF}-L}{(u_{ego} - u_{opp} + \frac{\delta U}{\Delta U}u_{boost}) cos\theta_{2}}
    \end{split}
    \label{eq:total_overtake_path}
\end{equation}

We consider the ego-racecar and the opponent-opponent racecar as a kinematic-bicycle model with the reference frame for both racecars at the center of the rear axle.
The overtake solution involves creating an energy efficient spline around the opponent-racecar from the ego-racecars current location.
We define energy efficiency in this context to be the amount of least additional energy that the ego-racecar can consume to perform an overtake.
We create a spline restricted by an envelope defined by four spline control points $G$ (guide).
Equation~\ref{eq:overtake_spline_control_points} provides an overview of how each point in $G$ is calculated, and a geometric description is provided in Figure~\ref{fig:quintic_planner_details} as the red spline.

\begin{equation}
    G =
    \begin{cases}
        g_0, & g(R_{ego}, W) \\
        g_1, & h(R_{opp}, g(R_{opp}, j(R_{opp}, B_L, B_R))) \\
        g_2, & h(k(R_{opp}, R_{opp}(\theta), L_{WB}), \\ 
        & g(k(R_{opp}, L_{WB}), j(R_{opp}, B_L, B_R))) \\
        g_3, & g(k(R_{opp}, L_{WB} + L_{CD}), W)
    \end{cases}
    \label{eq:overtake_spline_control_points}
\end{equation}

\begin{figure*}[h!]
    \centering
    \includegraphics[width=0.8\linewidth]{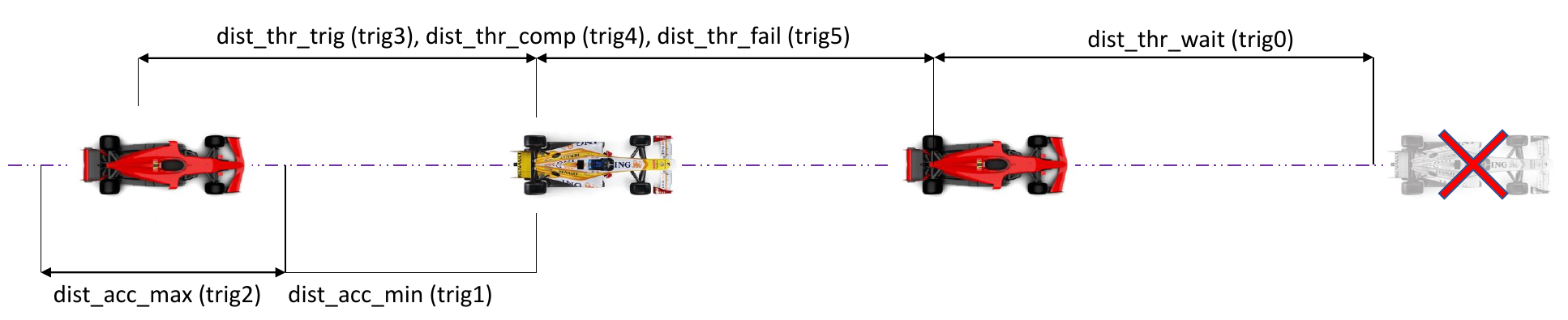}
    \caption{A geometric overview of the triggers in the ARGOS Framework where Red is the ego. The triggers \{trig3, trig4, trig5\} are distinct and are by design, close to each other in magnitude to allow dynamic corrections. The opponent car on the right marked with a cross is outside the tracking distance (trig0).}
    \label{fig:geometry_triggers}
\end{figure*}

\subsubsection{Modelling a Position Defense}

Similarly to overtaking, the position defense solution involves creating a local reference spline significantly different from the global raceline.
The primary objective of the position defense spline is to bring the ego-racecar close enough to the overtaking opponent-racecar and force the opponent-racecar to abandon the overtake.
Race requirement Rule \textbf{R3} places the responsibility of safeguarding safety distance on the attacker, therefore placing the ego-racecar close to the opponent-racecar will force the opponent-racecar to abandon the overtake.
Figure~\ref{fig:interceptor_geometry} describes the superprojection intercepting position-defense geometry used in this paper.
We used a superprojection analogous to the superelevation technique used in ballistic trajectory calculations.
In this figure, the projection of the opponent (i.e., the attacker) is the future pose of the attacker along its local overtake trajectory, and the super-projection is a multiple ($k$) of the racecar car length ($L$) added in front of the opponent's projected pose.
The defender is expected to find this superprojected pose and occupy it before the attacker in order to force the attacker to abandon the overtake.
The position defense spline, $I$ has three spline control (guide) points, each of which is described in Equation~\ref{eq:position_defense_spline_control_points}, and a geometric description is provided in Figure~\ref{fig:quintic_planner_details} as the green spline.

\begin{equation}
    I =
    \begin{cases}
        i_0, & g(R_{ego}, W) \\
        i_1, & h(k(R_{opp}, R_{opp}(\theta), R_{opp}(v)T_{p}), \\
        & g(k(R_{opp}, R_{opp}(v)T_{p}), j(R_{opp}, B_L, B_R))) \\
        i_2, & g(k(R_{opp}, R_{opp}(v)T_{p} + L_{CD}), W)
    \end{cases}
    \label{eq:position_defense_spline_control_points}
\end{equation}

In Section~\ref{sec:related_work}, we describe the stack used by a single racecar during an autonomous race.
The various nodes in the stack provide information to the racecar, but on its own, the decision to act on the information can be reactive, as shown in~\cite{IAC_1, IAC_2}.

In this paper, we present a formal solution to the H2H problem, where a network of interconnected state machines that, together with the perception, planning and control methods described in Section~\ref{sec:related_work}, will guide the racecar in making the correct decision during an overtake or position defense.
We choose the state machine approach as it helps us to easily decompose a complex maneuver such as an overtake to a temporal sequence of submaneuvers from which a racecar can safely progress through an attempted maneuver, or if it decides, safely abandon a maneuver.
More information about the submaneuvers is shown in Figure~\ref{fig:maneuver_trace_diagram}, with a timed odometry trace, that is, pose information about two racecars in a head-to-head autonomous race.
From this figure, the various events and submaneuvers involved in an overtake and position defense maneuvers are identified.
The two cases shown in Figure~\ref{fig:maneuver_trace_diagram} each depict opposite events.
In a head-to-head autonomous race, a successful overtake attempt by one racecar is a failed position defense by the other racecar; conversely, a successful position defense by one racecar is a failed overtake attempt by the other racecar.
Finally, a state machine approach also allows us to use formal methods to verify the functioning of the proposed framework.

Triggers are the hard constraints that the components of the ARGOS framework use to initiate an event.
These values can be tuned to get the desired results, and they form the basis of measuring and validating traces during model checking and design verification.
Table~\ref{tab:argos_triggers} provides a summary of the triggers of the ARGOS framework.

\begin{itemize}
    \item Opponent Tracking Distance (trig0):
    The ego will track the opponent racecar only up to a certain Euclidean distance.
    This is largely due to the sensor-ranging limitations, but also to optimize implementation.
    
    \item Minimum Follow Distance (trig1) and Maximum Follow Distance (trig2):
    The closed range of distances defined by these triggers describe the distance behind the opponent that the ego racecar must maintain when a passing maneuver is not possible.
    
    \item Minimum Distance to Trigger Maneuver Start (trig3):
    The minimum distance from the opponent that the ego racecar has to maintain to perform a maneuver.
    
    \item Minimum Distance to Trigger Maneuver Complete (trig4):
    When the ego racecar has successfully completed a maneuver, it must reach a safe distance in front of the opponent in order to merge back on the raceline.
    
    \item Minimum Distance to Trigger Maneuver Failure (trig5):
    If the ego racecar did not successfully complete a maneuver and has to abandon or fallback, it must clear a safe distance from the opponent to resume the race.
    
    \item AEMS Trigger Warning (trig6) and AEMS Trigger Failure (trig7):
    These triggers inform the ARGOS Framework before (trig6) an overtake attempt if the requested time-energy budget is available in the AEMS reservoir and during (trig7) an overtake attempt if the ego should abandon due to the lack of energy budget.
\end{itemize}

\begin{table}[h!]
\centering
\begin{tabular}{|l|l|l|}
\hline
Symbol & Description \\ \hline
trig0 & Max opponent tracking dist \\ \hline
trig1 & Min follow dist for tracking \\ \hline
trig2 & Max follow dist for tracking \\ \hline
trig3 & Min dist to trigger a pass or block \\ \hline
trig4 & Min dist to trigger a maneuver complete \\ \hline
trig5 & Min dist to recover from a failed maneuver \\ \hline
trig6 & Min pre-start maneuver budget \\ \hline
trig7 & Min post-start maneuver budget \\ \hline
trig8 & Min lateral separation between racecars \\ \hline
\end{tabular}
\caption{A summary list of thresholds and triggers used in the ARGOS framework}
\label{tab:argos_triggers}
\end{table}

\begin{figure*}[h!]
    \centering
    \includegraphics[width=0.8\linewidth]{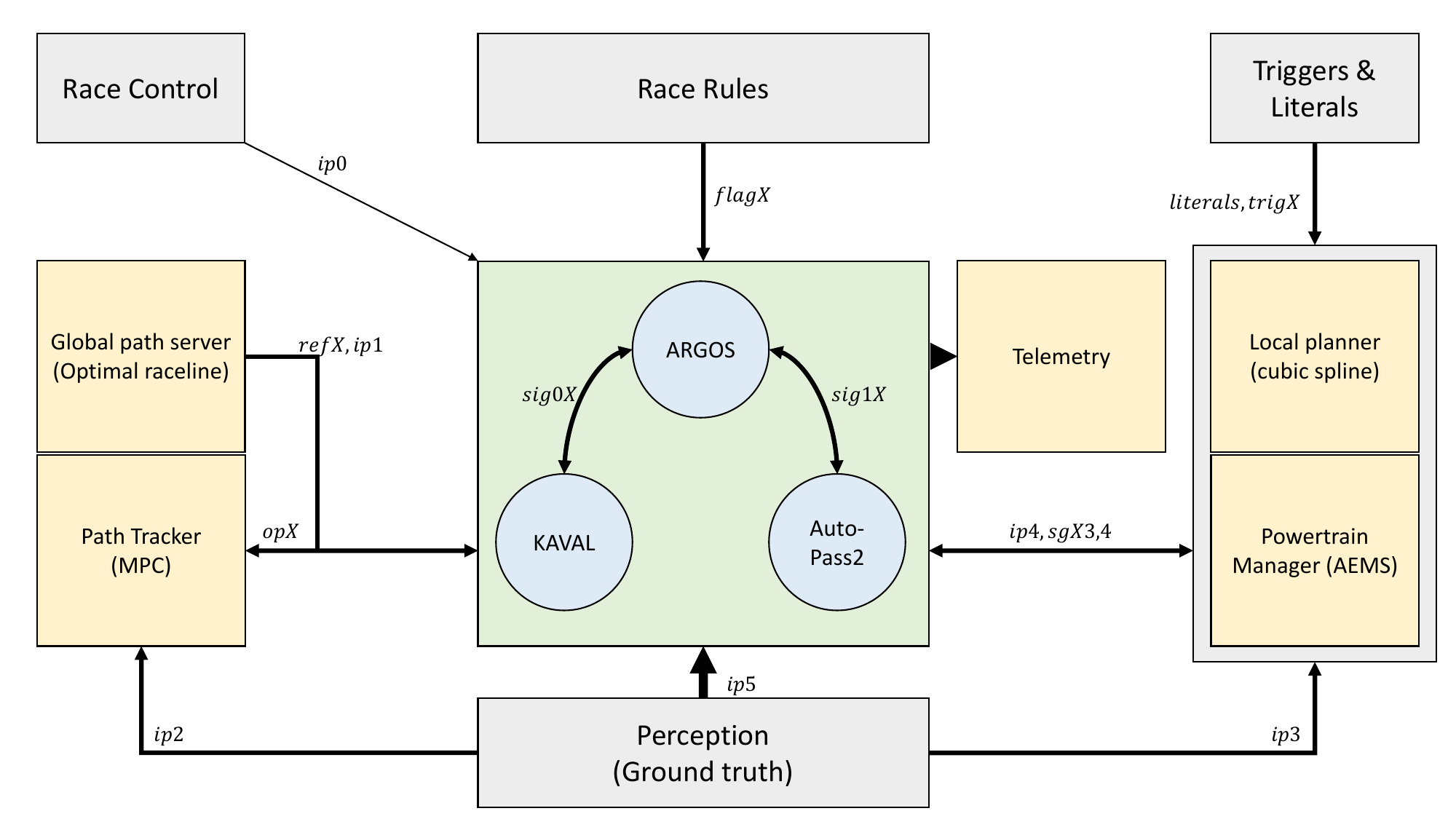}
    \caption{An overview of the ARGOS framework architecture with intra-framework I/O. Gray blocks are the external signals, rules and triggers, Yellow blocks are the autonomous racing support nodes, and Green is the ARGOS framework with the three automatons (ARGOS, AutoPass. and KAVAL) shown in Blue.}
    \label{fig:argos_arch}
\end{figure*}

\section{ARGOS Framework}
\label{sec:ARGOS_framework}
In this section, we describe the Automaton Referencing Guided Overtake System (ARGOS) framework and automaton, the 2\textsuperscript{nd} iteration of the Autonomous Passing framework (AutoPass), and the Kinematic Vector Analysis (KAVAL) position defense framework.

\subsection{ARGOS Architecture}

The ARGOS framework architecture is presented in Figure~\ref{fig:argos_arch}.
The framework consists of 3 interconnected automatons: ARGOS, AutoPass and KAVAL, a set of hard rules encoded as triggers, some static resources for reference (eg: a global optimal raceline, and velocity profile, etc.) that work along with a set of support nodes to enable fully autonomous head-to-head racing.
These nodes are described in detail in Section~\ref{sec:related_work}, but a brief summary is presented here.

The support nodes include (a) the global path server, (b) the Model Predictive Control (MPC) path tracker, and (c) the race control module.
The framework nodes include (a) the cubic spline local planner, (b) power-train manager with the Advanced Energy Management System (AEMS), and (c) the state estimation module (in this paper, this is the simulator's ground truth relay node).
The automaton nodes include (a) the ARGOS automaton, (b) the AutoPass automaton, and (c) the KAVAL automaton and their associated signals and triggers.

\subsection{Framework I/O}

\noindent \textbf{Inputs:} The ARGOS framework uses the following states as input.
Table~\ref{tab:argos_inputs} provides a summary of the ARGOS Framework Inputs.

\begin{itemize}
    \item Race flag (\textbf{ip0}):
    The race flag is set by the Race Control module.
    The race flag describes the condition of the track and the expected behaviors of all agents on the track.
    The race flag also sets performance constraints such as (a) velocity limits, (b) enables maneuvers like overtaking and position defense, etc.
    
    \item Reference trajectory (\textbf{ip1}):
    The reference trajectory is a set of local waypoints within the track-bounds joined using a spline.
    The waypoints are encoded with a desired velocity component similar to the global raceline.
    This spline is used by the Path Tracker to move the racecar along the reference trajectory.
    
    \item Longitudinal Separation (\textbf{ip2}):
    Longitudinal separation is the length of the global race line that separates the ego-racecar and the opponent-racecar.
    It is a measure of the true geometric separation between the racecars when determining their kinematic constraints.
    
    \item Leader State (\textbf{ip3}):
    The leader state is a flag that informs the ARGOS framework if the ego-racecar is the current race leader.
    ARGOS uses this information to determine the role of the ego-racecar as an attacker or a defender.
    
    \item AEMS Reservoir State (\textbf{ip4}):
    The AEMS reservoir state is the current available boost energy budget available to the ego-racecar.
    ARGOS uses this information to determine whether a planned maneuver is likely to be successful.
    
    \item Opponent State (\textbf{ip5}):
    The opponent state is a one-hot vector of four elements (\textbf{OHV}) that provides an estimation of the opponent-racecars' intentions.
    The four elements are in the following order:
    \begin{itemize}
        \item opponent is attempting to overtake
        \item opponent has abandoned the overtake
        \item opponent is attempting to block an overtake
        \item opponent has decided to fallback
    \end{itemize}
    The vector is decomposed as a Binary Coded Decimal in the form $OHV \langle X \rangle$.
\end{itemize}

\begin{table}[h!]
\centering
\begin{tabular}{|l|l|l|}
\hline
Symbol & Description \\ \hline
ip0 & Current race control flag \\ \hline
ip1 & The active path tracker trajectory \\ \hline
ip2 & The path separation between ego and opponent \\ \hline
ip3 & Flag that is set when ego is the race leader \\ \hline
ip4 & The remaining AEMS time-energy budget \\ \hline
ip5 & One Hot Vector (OHV) of the opponent state \\ \hline
\end{tabular}
\caption{A summary list of inputs and their notation used by the ARGOS framework}
\label{tab:argos_inputs}
\end{table}

\noindent \textbf{Signals:} Signals are the internal framework states.
The components of the framework have well-defined rules to modify the internal framework states.
Table~\ref{tab:argos_signals} provides a summary of the ARGOS Framework signals.

\begin{itemize}
    \item AutoPass Arm (sg00) and KAVAL Arm (sg10):
    This signal sends the arm command to AutoPass or KAVAL.
    Arming is necessary to allow the local overtake planner and the AEMS node.
    
    \item AutoPass Initiate (sg01) and KAVAL Initiate (sg11):
    This signal allows AutoPass or KAVAL to produce local reference trajectories.
    ARGOS will use the local reference trajectory from this state until a framework reset occurs, either through a successful or failed maneuver.
    
    \item AutoPass Maneuver Complete (sg02) and KAVAL Maneuver Complete (sg12):
    This is the reset signal that informs ARGOS about a completed maneuver (either successful or failed).
    
    \item AutoPass Velocity Override (sg03) and AutoPass Trajectory Override (sg04):
    When AutoPass controls the ego racecar, these signals provide local velocity and trajectory references to the Path Tracker.
    
    \item KAVAL Velocity Override (sg13) and KAVAL Trajectory Override (sg14):
    When KAVAL controls the ego racecar, these signals provide local velocity and trajectory references to the Path Tracker.
\end{itemize}

\begin{table}[h!]
\centering
\begin{tabular}{|l|l|l|}
\hline
Symbol & Description \\ \hline
sg00 & AutoPass Arm signal \\ \hline
sg01 & AutoPass Init signal \\ \hline
sg02 & AutoPass Maneuver complete signal \\ \hline
sg03 & AutoPass Velocity override signal \\ \hline
sg04 & AutoPass Reference path override signal \\ \hline
sg10 & KAVAL Arm signal \\ \hline
sg11 & KAVAL Init signal \\ \hline
sg12 & KAVAL Maneuver complete signal \\ \hline
sg13 & KAVAL Velocity override signal \\ \hline
sg14 & KAVAL Reference path override signal \\ \hline
sg90 & Global reference velocity profile \\ \hline
sg91 & Global reference trajectory \\ \hline
\end{tabular}
\caption{A summary list of signals and notation within the ARGOS framework.}
\label{tab:argos_signals}
\end{table}

\noindent \textbf{Outputs:} The ARGOS framework outputs a reference trajectory and an associated velocity profile. 
These outputs are multiplexed using the One Hot Vector (OHV) bit mask.
Table~\ref{tab:argos_outputs} provides a summary of the ARGOS framework outputs.

\begin{itemize}
    \item Output Master (op1): A flag informs the path tracker that an override is necessary.
    \item Velocity Override Reference (op2) and Trajectory Override Reference (op3):
    Based on the OHV, the ARGOS Framework sends the desired override velocity profile and local reference trajectory to the path tracker.
\end{itemize}

\begin{table}[h!]
\centering
\begin{tabular}{|l|l|l|}
\hline
Symbol & Description \\ \hline
op0 & OHV of the override control output \\ \hline
op1 & The output target velocity for path tracker \\ \hline
op2 & The output target trajectory for path tracker \\ \hline
\end{tabular}
\caption{A summary list of outputs and their notations used by the ARGOS framework}
\label{tab:argos_outputs}
\end{table}

\subsection{ARGOS Automaton}

\begin{figure}[h!]
\centering
\includegraphics[width=\columnwidth]{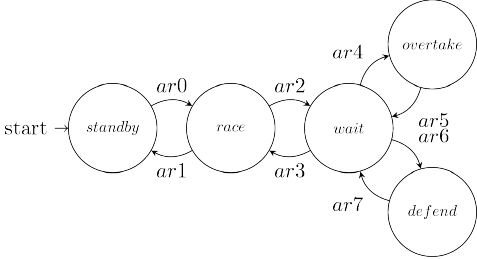}
\caption{The ARGOS automaton.}
\label{fig:argos_automaton}
\end{figure}

The ARGOS automaton is the central automaton that supervises the framework.
The ARGOS automaton informs the racecar about the race rules and any potential violations.
This section provides a brief description of the ARGOS automaton.
The automaton is represented in Figure~\ref{fig:argos_automaton} that shows the states and inter-state transition conditions.


    


The states of the ARGOS automaton are:
(1) \textbf{Standby}, during which the entire ARGOS framework is disabled and the racecar is tracking the trajectory to the pit box via the pit-lane trajectory connecting the race track to the pit box, 
(2) \textbf{Race}, during which the ego racecar is following the global raceline and the opponent is not within the monitoring range,
(3) \textbf{Wait}, during which the ego continues to follow the global raceline with the opponent in the monitoring range and the ego arms either the AutoPass (for overtake, if the opponent is the race leader), or KAVAL (for position defense, if ego is the race leader).
\textbf{Overtake}, when the ego racecar is cleared for overtake by race control, and has a viable overtake trajectory and boost-energy budget, the ARGOS automaton initiates the AutoPass automaton and remains in this state until either an overtake is completed or abandoned,
(4) \textbf{Defend}, when the ego racecar is cleared for a position defense, and it has a viable trajectory to block the opponent and the ego estimates a high probability of success in blocking the opponent, when the ARGOS automaton initiates the KAVAL automaton and remains in this state until the position defense maneuver is completed or the opponent has managed to pass the ego racecar.

\begin{equation}
    G_{ar} = \begin{cases}
        ar0: & ip1 \langle raceline \rangle \\
        ar1: & ip0 \langle Black \rangle \\
        ar2: & ip2 \in [trig1, trig2] \\
        ar3: & ip2 \notin [trig1, trig2] \\
        ar4: & \sim ip3 \wedge ip0 \langle Blue \rangle \\
        ar5: & sg02 \\
        ar6: & ip3 \wedge ip0 \langle Blue \rangle \\
        ar7: & sg12 
    \end{cases}
    \label{eq:argos_guard_cond}
\end{equation}

\begin{figure}[h!]
    \centering
    \includegraphics[width=2\linewidth, angle=270]{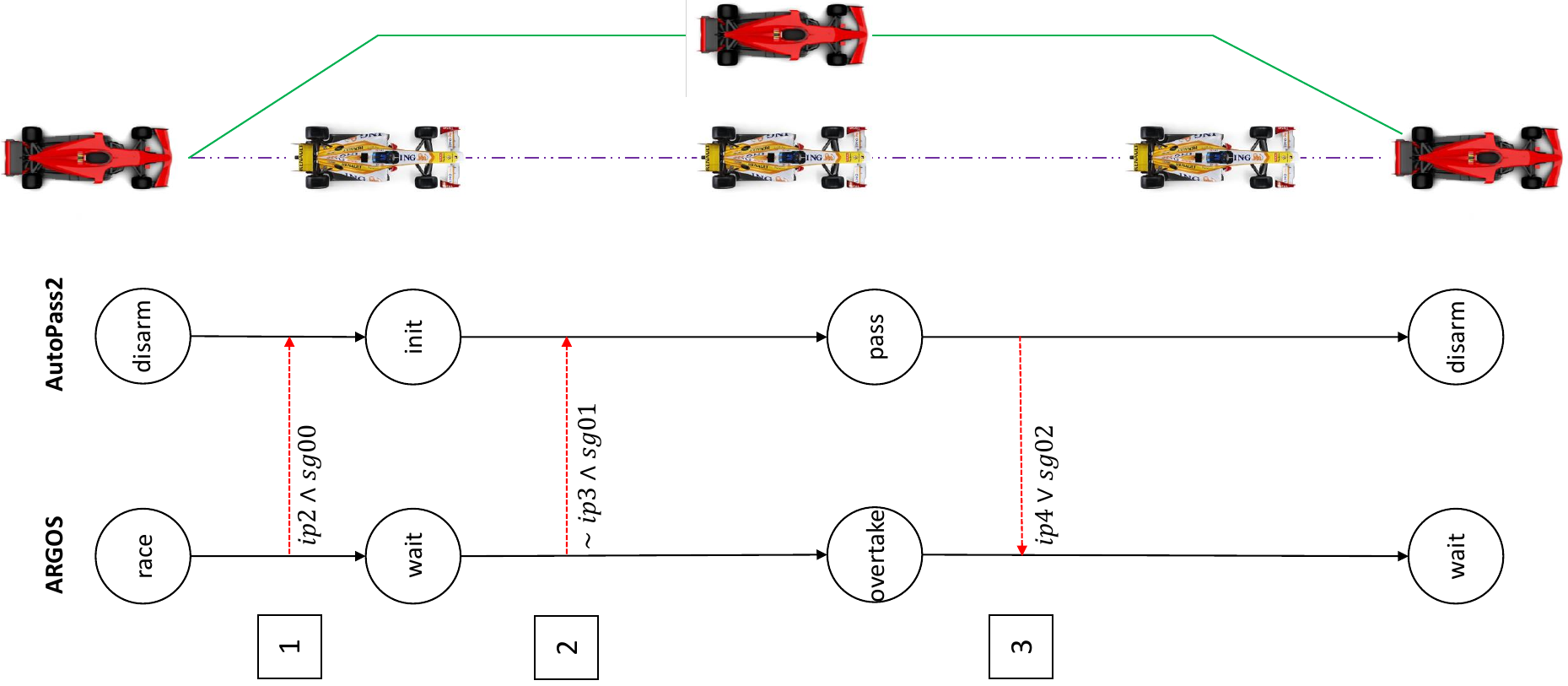}
    \caption{
        A geometric overview of an overtake by the ARGOS framework. Green and red are ego and opponent trajectory respectively.
        In numeric order from the figure: 
        (1) ARGOS arms AutoPass using \textbf{sg00},
        (2) ARGOS initiates an overtake using \textbf{sg01} and hands over control to AutoPass,
        (3) AutoPass completes an overtake and sets \textbf{sg03}, informing ARGOS that the overtake maneuver is complete, and resets and disarms AutoPass.}
    \label{fig:AutoPass_geometry}
\end{figure}

In Equation~\ref{eq:argos_guard_cond}, $\langle Blue \rangle$ flag indicates that the racecar is in the passing zone, and the $\langle Black \rangle$ flag triggers the end of the race.
While the overtake and position defense trajectories are generated using the cubic spline planner described later in this section, and the boost-energy module independently provides and monitors the boost-energy budget, the ARGOS automaton, when in the overtake or defense state, sets the framework output \textbf{op0} using Equation~\ref{eq:argos_ohv_set}.
When set, in Big-Endian, the first two positions in this output represent the request from AutoPass (to use local planner trajectory and speed profile, respectively) and the last two represent the request from KAVAL.

\begin{equation}
    op0 = \begin{cases}
        OHV \langle 0 \rangle: & \sim sg03 \vee sg04 \vee sg13 \vee sg14 \\
        OHV \langle 1 \rangle: & \sim sg03 \vee sg13 \vee sg14 \wedge sg04 \\
        OHV \langle 2 \rangle: & \sim sg04 \vee sg13 \vee sg 14 \wedge sg03 \\
        OHV \langle 3 \rangle: & \sim sg13 \vee sg14 \wedge sg03 \wedge sg04 \\
        OHV \langle 4 \rangle: & \sim sg03 \vee sg04 \vee sg13 \wedge sg14 \\
        OHV \langle 8 \rangle: & \sim sg03 \vee sg04 \vee sg14 \wedge sg13 \\
        OHV \langle 12 \rangle: & \sim sg03 \vee sg04 \wedge sg13 \wedge sg14
    \end{cases}
\label{eq:argos_ohv_set}
\end{equation}

Equations~\ref{eq:argos_vel_od} and~\ref{eq:argos_path_od} describe the velocity profile and the active reference trajectory, respectively, being used by the path tracker.
When ARGOS decides during an overtake or position defense that the target velocity or the reference trajectory needs to be changed, it will use the corresponding active override signal to make the path tracker follow the new reference.

\begin{equation}
    op1 = \begin{cases}
        sg90: & op0 \langle 0 \rangle \\
        sg03: & op0 \langle 2 \rangle \vee op0 \langle 3 \rangle \\
        sg13: & op0 \langle 8 \rangle \vee op0 \langle 12 \rangle
    \end{cases}
\label{eq:argos_vel_od}
\end{equation}

\begin{equation}
    op2 = \begin{cases}
        sg91: & op0 \langle 0 \rangle \\
        sg04: & op0 \langle 1 \rangle \vee op0 \langle 3 \rangle \\
        sg14: & op0 \langle 4 \rangle \vee op0 \langle 12 \rangle
    \end{cases}
\label{eq:argos_path_od}
\end{equation}

\subsection{AutoPass Automaton}

\begin{figure}[h!]
\centering
\includegraphics[width=\columnwidth]{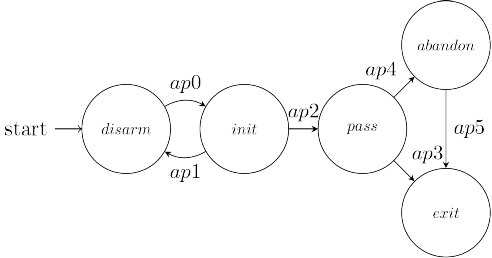}
\caption{The AutoPass automaton.}
\label{fig:AutoPass_automaton}
\end{figure}

The AutoPass automaton, referenced in Figure~\ref{fig:AutoPass_automaton}, is designed to inform the ego racecar if it can safely pass the opponent and, if necessary, AutoPass will safely abandon the overtake attempt and return to the global raceline behind the opponent.


    


The states of the AutoPass automaton are: 
(1) \textbf{Disarm}, during which AutoPass and its internal states are continuously reset,
(2) \textbf{Init}, during which, AutoPass is armed, and the local planner generated overtake trajectory is monitored for feasibility and safety constraints imposed by the triggers defined in the previous section, and AutoPass remains in this state if an overtake is not possible before exiting,
(3) \textbf{Pass}, when ARGOS requests an overtake, AutoPass draws energy from the boost energy budget and tracks the speed profile generated by the planner when the overtake trajectory is feasible,
(4) \textbf{Abandon}, when during an overtake attempt, the opponent successfully blocks the ego, or if the ego has depleted the boost energy budget, the ego slows down and merges on to the global raceline at a safe distance behind the opponent as defined in \textbf{trig5}, 
(5) \textbf{Exit} is a transient state that informs ARGOS that the overtake maneuver is complete and resets AutoPass and the local planner.

\begin{equation}
    G_{ap} = \begin{cases}
        ap0: & sg00 \\
        ap1: & \sim sg00 \\
        ap2: & sg01 \\
        ap3: & ip4 > trig6 \wedge ip5 \langle 2 \rangle \\
        ap4: & ip4 < trig7 \vee \sim ip5 \langle 2 \rangle \\
        ap5: & sg02 \wedge ip2 \in [trig1, trig2]
    \end{cases}
\label{eq:AutoPass_guard_cond}
\end{equation}

Equation~\ref{eq:AutoPass_guard_cond} provides the logical description of all transition conditions in the AutoPass automaton shown in Figure~\ref{fig:AutoPass_automaton}.
The signals, triggers, and inputs are defined in the previous sections. The dynamics of the individual states do not affect the ego racecar unless AutoPass is in Pass or Abandon.
Equations~\ref{eq:AutoPass_vel_od} and~\ref{eq:AutoPass_path_od} describe the local velocity profile and reference trajectory provided by AutoPass.

\begin{equation}
    op1_{ap} = \begin{cases}
        disarm: & sg90 \\
        init: & sg90 \\
        pass: & sg03 \\
        abandon: & sg03 \longrightarrow ap4 \\
        exit: & sg90
    \end{cases}
\label{eq:AutoPass_vel_od}
\end{equation}

\begin{equation}
    op2_{ap} = \begin{cases}
        disarm: & sg91 \\
        init: & sg91 \\
        pass: & sg04 \\
        abandon: & sg04 \longrightarrow ap4 \\
        exit: & sg91
    \end{cases}
\label{eq:AutoPass_path_od}
\end{equation}

\subsection{KAVAL Automaton}

\begin{figure}[h!]
\centering
\includegraphics[width=\columnwidth]{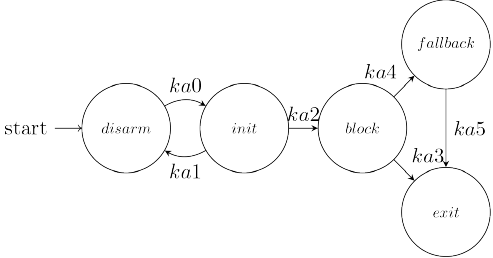}
\caption{The KAVAL automaton.}
\label{fig:kaval_automaton}
\end{figure}

The KAVAL automaton, referenced in Figure~\ref{fig:kaval_automaton}, is designed to inform the ego racecar if and when it can safely block the opponent for an attempted overtake, and if the position defense maneuver is unsuccessful (the opponent racecar has managed to overcome the block attempt), KAVAL will safely abandon the block attempt and fallback on to the global raceline behind the opponent.


    


\begin{figure}[h!]
    \centering
    \includegraphics[width=2\linewidth, angle=270]{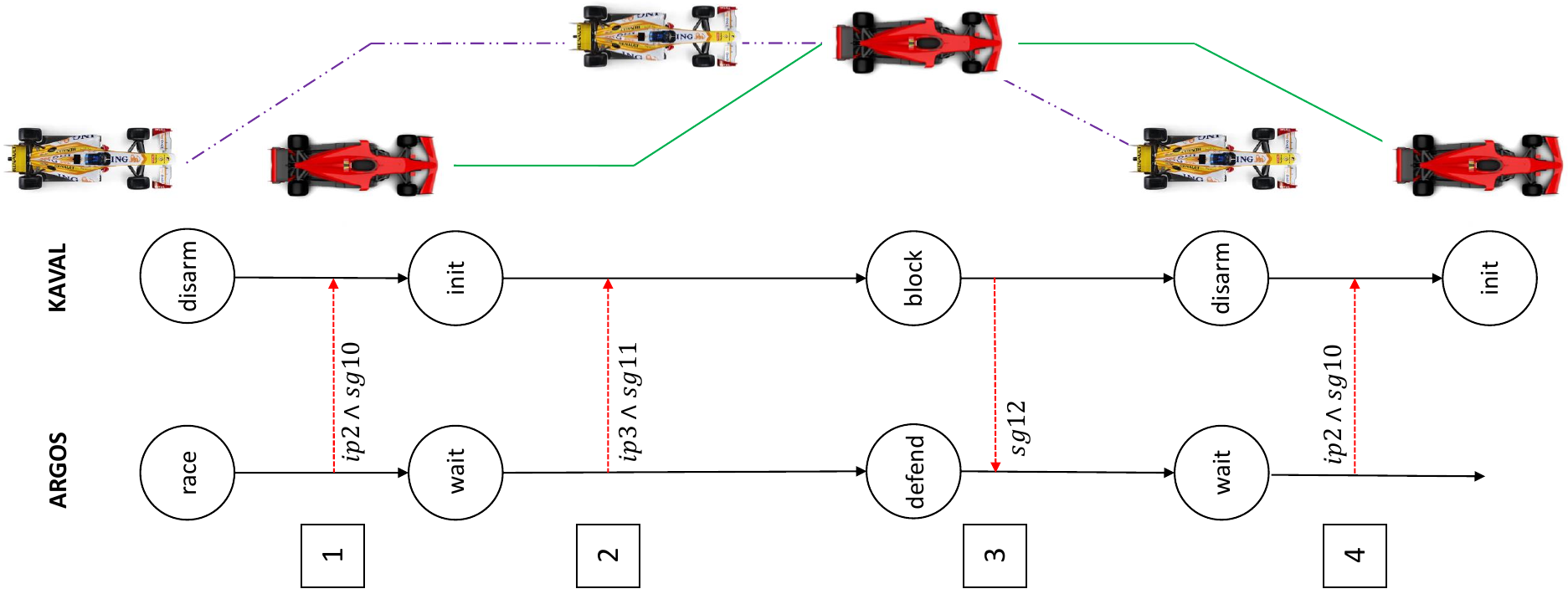}
    \caption{
        A geometric overview of a successful position defense by the ARGOS framework. Green and red are ego and opponent trajectory respectively.
        In numeric order from the figure: 
        (1) ARGOS arms KAVAL using \textbf{sg10},
        (2) ARGOS initiates a position defense using \textbf{s101} and hands over control to KAVAL,
        (3) KAVAL completes the defense maneuver and sets \textbf{s13}, informing ARGOS that the overtake maneuver is complete, and resets and disarms KAVAL,
        (4) Since ego retains the race leader position, ARGOS arms KAVAL using \textbf{sg10}.}
    \label{fig:kaval_geometry}
\end{figure}

The states of the KAVAL automaton are: 
(1) \textbf{Disarm}, during which KAVAL and its internal states are continuously reset,
(2) \textbf{Init}, during which KAVAL is armed, and the local planner generated position defense trajectory is monitored for feasibility and safety constraints imposed by the triggers defined in the previous section, and KAVAL remains in this state if the opponent does not attempt to pass or if a position defense is not feasible,
(3) \textbf{Block}, when ARGOS requests a block attempt for position defense, KAVAL informs the ego racecar to track the trajectory generated by the local planner that will intercept the current motion vector of the opponent,
(4) \textbf{Fallback}, if the position defense attempt is unsuccessful, KAVAL slows down and merges on to the global raceline at a safe distance behind the opponent,
(5) \textbf{Exit} is a transient state that informs ARGOS that the position defense maneuver is complete, and resets KAVAL and the local planner.

Equation~\ref{eq:kaval_guard_cond} provides the logical description of all transition conditions in the KAVAL automaton shown in Figure~\ref{fig:kaval_automaton}.
The signals, triggers and inputs are defined in the previous sections.
The dynamics of the individual states do not affect the ego racecar unless AutoPass2 is in Block or Fallback.
Equations~\ref{eq:kaval_vel_od} and~\ref{eq:kaval_path_od} describe the local velocity profile and reference trajectory provided by KAVAL.

\begin{equation}
    G_{ka} = \begin{cases}
        ka0: & sg10 \\
        ka1: & \sim sg10 \\
        ka2: & sg11 \\
        ka3: & sg12 \wedge ip5 \langle 8 \rangle \\
        ka4: & sg12 \wedge \sim ip5 \langle 8 \rangle \\
        ka5: & sg12 \wedge ip2 \in [trig0, trig1]
    \end{cases}
\label{eq:kaval_guard_cond}
\end{equation}

\begin{equation}
    op1_{ka} = \begin{cases}
        disarm: & sg90 \\
        init: & sg90 \\
        pass: & sg13 \\
        abandon: & \{sg13 \longrightarrow ka4\} \\
        exit: & sg90
    \end{cases}
\label{eq:kaval_vel_od}
\end{equation}

\begin{equation}
    op2_{ka} = \begin{cases}
        disarm: & sg91 \\
        init: & sg91 \\
        pass: & sg14 \\
        abandon: & sg14 \longrightarrow ka4 \\
        exit: & sg91
    \end{cases}
\label{eq:kaval_path_od}
\end{equation}

\begin{figure*}[h!]
    \centering
    \includegraphics[width=\linewidth]{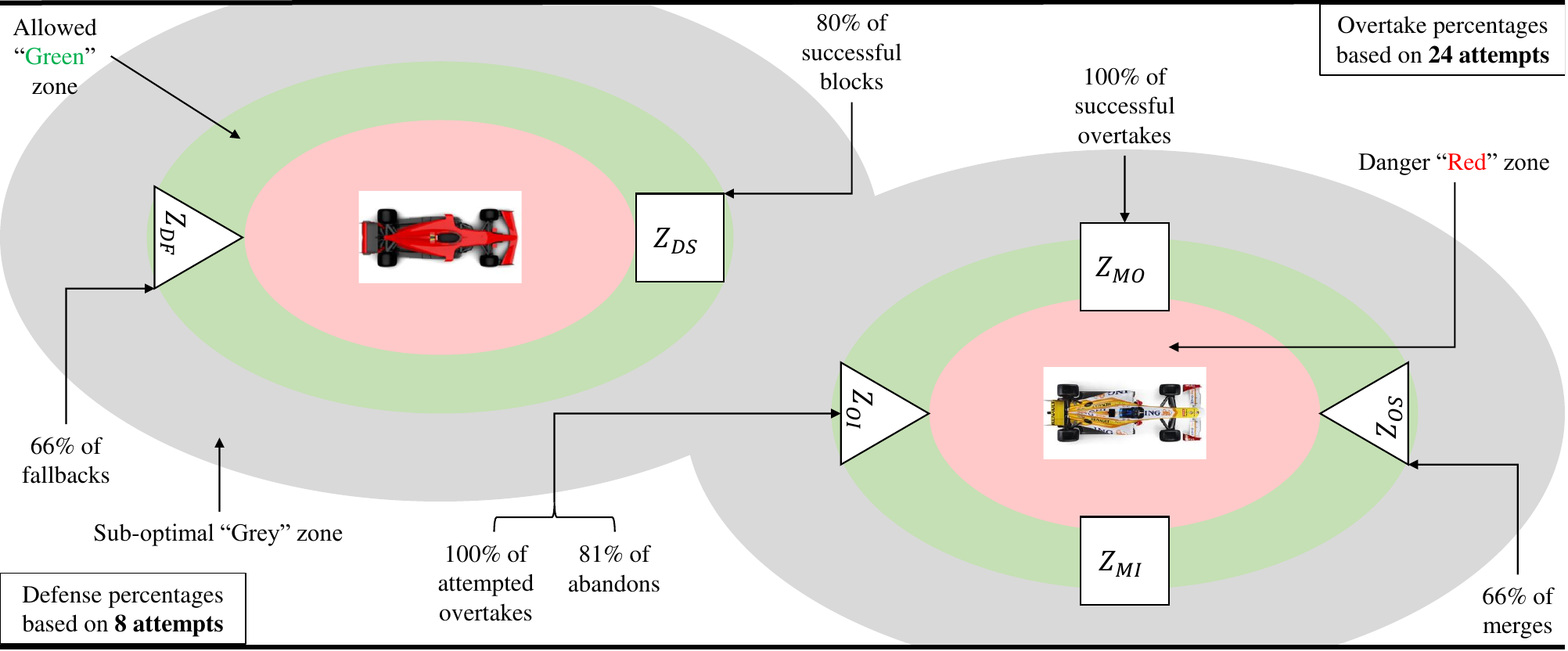}
    \caption{A geometric threshold compliance diagram showing the position of the overtaking racecar (Red) and the defending racecar (Yellow), and the accompanying zone classifications shown in Red (danger), Green (allowed), and Grey (sub-optimal). Numerical percentages shown here are from the experimental results shown in Figure~\ref{fig:trace_combo_result}.}
    \label{fig:geometric_threshold_result}
\end{figure*}

\section{Formal Verification}
\label{sec:model_checking}
On the surface, the ARGOS framework appears complete and up to the standards specified in Section~\ref{sec:h2h_problem}, where we provide an overview of the requirements and expected behavior of the ARGOS framework.
However, to verify that the ARGOS framework performed as designed continuously in a deterministic manner, we used formal methods, specifically design verification and model verification.
We used Matlab's Formal Verification toolbox to perform the tasks in this section.

\subsection{Framework Design Verification}

Design verification is the first step in the formal verification process.
The principal task here is to identify dead-logic and unreachable states.
Dead-logic is defined as any guard condition within the ARGOS automaton network that is never triggered, and an unreachable state is that target state of the corresponding dead logic.
An ideal automaton-based framework will be designed to trigger a transition for every event, and the transition can be to the same state (self-transition) or to a different state (outtransition).
Design verification involves formally specifying the requirements of the framework (pre-design specification) and creating a valid list of temporal state sequences (framework temporal logic).

\noindent \textbf{Pre-Design Specifications}
In this step, we verify the transition-in and transition-out properties of each state within the ARGOS framework.
Each state in each automaton within the framework must have unique guard conditions, and each automaton can have only one active state at any time.
The ARGOS framework consists of unique guard conditions as defined in Equations~\ref{eq:argos_guard_cond},\ref{eq:AutoPass_guard_cond} and~\ref{eq:kaval_guard_cond}, with each guard condition referencing well-defined triggers in Table~\ref{tab:argos_triggers}.

\noindent \textbf{Framework Temporal Logic}

\begin{table}[h!]
    \centering
        \begin{tabular}{|l|l|l|l|}
        \hline
        ARGOS & AutoPass & KAVAL & FSC tag \\ \hline
        standby & disarm & disarm & \textit{fsc00} \\ \hline
        race & disarm & disarm & \textit{fsc10} \\ \hline
        wait & init & disarm & \textit{fsc20} \\ \hline
        wait & disarm & init & \textit{fsc21} \\ \hline
        overtake & pass & disarm & \textit{fsc30} \\ \hline
        overtake & abandon & disarm & \textit{fsc31} \\ \hline
        defend & disarm & block & \textit{fsc40} \\ \hline
        defend & disarm & fallback & \textit{fsc41} \\ \hline
        \end{tabular}
    \caption{A list of valid Framework State Combinations (FSC) for the fully integrated ARGOS framework}
    \label{tab:state_combinations}
\end{table}

The state machines together in the ARGOS framework are allowed to exist in one unique combination of states, called the Framework State Combination (FSC).
All valid FSCs use the tag $fscXX$ and are listed in Table~\ref{tab:state_combinations}.
The framework temporal logic is the temporal sequence makeup of the FSCs that make up a defined maneuver.
    Within the ARGOS framework, we define the following temporal sequences:
\begin{itemize}
    \item \textbf{Successful Overtake}: A successful overtake is made up of two concatenated events, and each event has an associated temporal sequence: (a) an overtake initiation ($fsc10-fsc20-fsc30$), and (b) a successful overtake ($fsc30-fsc20-fsc10$)
    \item \textbf{Abandoned Overtake}: A failed overtake is made up of three concatenated events: (a) an initiation of the overtake ($fsc10-fsc20-fsc30$), (b) an failed overtake trigger ($fsc30-fsc31$), and (c) a successful abandon ($fsc31-fsc20-fsc10$)
    \item \textbf{Successful Defense}: A successful position defense is made up of two concatenated events: (a) a defense initiation ($fsc10-fsc21-fsc40$), and (b) a successful defense ($fsc40-fsc21-fsc10$)
    \item \textbf{Failed Defense}: A failed position is made up of three concatenated events: (a) a defense initiation ($fsc10-fsc21-fsc40$), (b) a position defense failed trigger, usually indicated by a change in race reader state ($fsc40-fsc41$), and (c) a successful fallback ($fsc41-fsc21-fsc10$)
\end{itemize}

For the FSCs defined in Table~\ref{tab:state_combinations}, the temporal logic for the ARGOS framework is summarized in Equation~\ref{eq:argos_ltl}.

\begin{equation}
    E_\varphi = \begin{cases}
        e_0: & G (fsc10) U (fsc20) U (fsc30) \\
             & X (fsc21) U (fsc10) \\
        e_1: & G (fsc10) U (fsc20) U (fsc30) \\
             & X (fsc31) X (fsc20) U (fsc10) \\
        e_2: & G (fsc10) U (fsc21) U (fsc40) \\
             & X (fsc20) U (fsc10) \\
        e_3: & G (fsc10) U (fsc21) U (fsc40) \\
             & X (fsc41) X (fsc21) U (fsc10)
    \end{cases}
\label{eq:argos_ltl}
\end{equation}

\subsection{Framework Model-Checking}

Once we verified that the ARGOS framework had well-defined requirements and a automaton network free of dead-logic and unreachable states, we proceeded to perform model-checking of the ARGOS framework.

\noindent \textbf{Checking against Framework Specification}

Model-checking against three automatons simultaneously is computationally expensive.
Within the ARGOS framework, the state transition sequences are cyclical, but infinite - this leads to the state explosion problem mentioned in~\cite{model_checking_1} and described in detail in~\cite{ model_checking_2, model_checking_3}.
One possible workaround is to simplify the \textbf{Overtake} and \textbf{Defense} states of the ARGOS automaton by using a truth table derived from independent model checks of AutoPass and KAVAL automatons, respectively.
Now, we define the ARGOS automaton as $M \langle S, S_0, \delta, F \rangle$, following the definitions used in~\cite{model_checking_1} with a set $E_M$ of all possible traces within the automaton.
To be considered a complete model, the traces $E_\varphi$ defined in Equation~\ref{eq:argos_ltl} must be a subset of $E_M$, that is, $E_M \subset E_\varphi$.

\noindent \textbf{Redesign using Counter-Examples}
If $E_M \notin E_\varphi$, the model checker tool (in our case, the Matlab design verifier) would produce a counterexample to indicate which sequence in $E_M$ is outside of $E_\varphi$.
This information is used to redesign the logic of the failing guard condition, and the process is repeated in an iterative manner until the model satisfies the requirements of Equation~\ref{eq:argos_ltl}.
While the ARGOS framework presented in the paper is in its complete form, we encountered many counterexamples that failed the requirements defined in $E_\varphi$.
An instance where a counterexample helped refine the ARGOS framework was AutoPass guard condition $ap4$.
Originally, the logic was $ap4 = ip4 < trig7$, which would provide a false positive condition where the ego-racecars' boost energy budget falling under $trig7$ would automatically trigger an overtake abandon, but if the opponent-racecar abandoned a position defense, the ego would still have the opportunity to complete an overtake.
This would lead to an invalid FSC as defined in Table~\ref{tab:state_combinations} where AutoPass is in the abandon state and KAVAL is in the fallback state.
To keep AutoPass in the pass state when KAVAL is in the abandon state, we modify the guard condition to $ap4 = ip4 < trig7 \vee \sim ip5 \langle 2 \rangle$.

\begin{figure*}[h!]
    \centering
    \includegraphics[width=\linewidth]{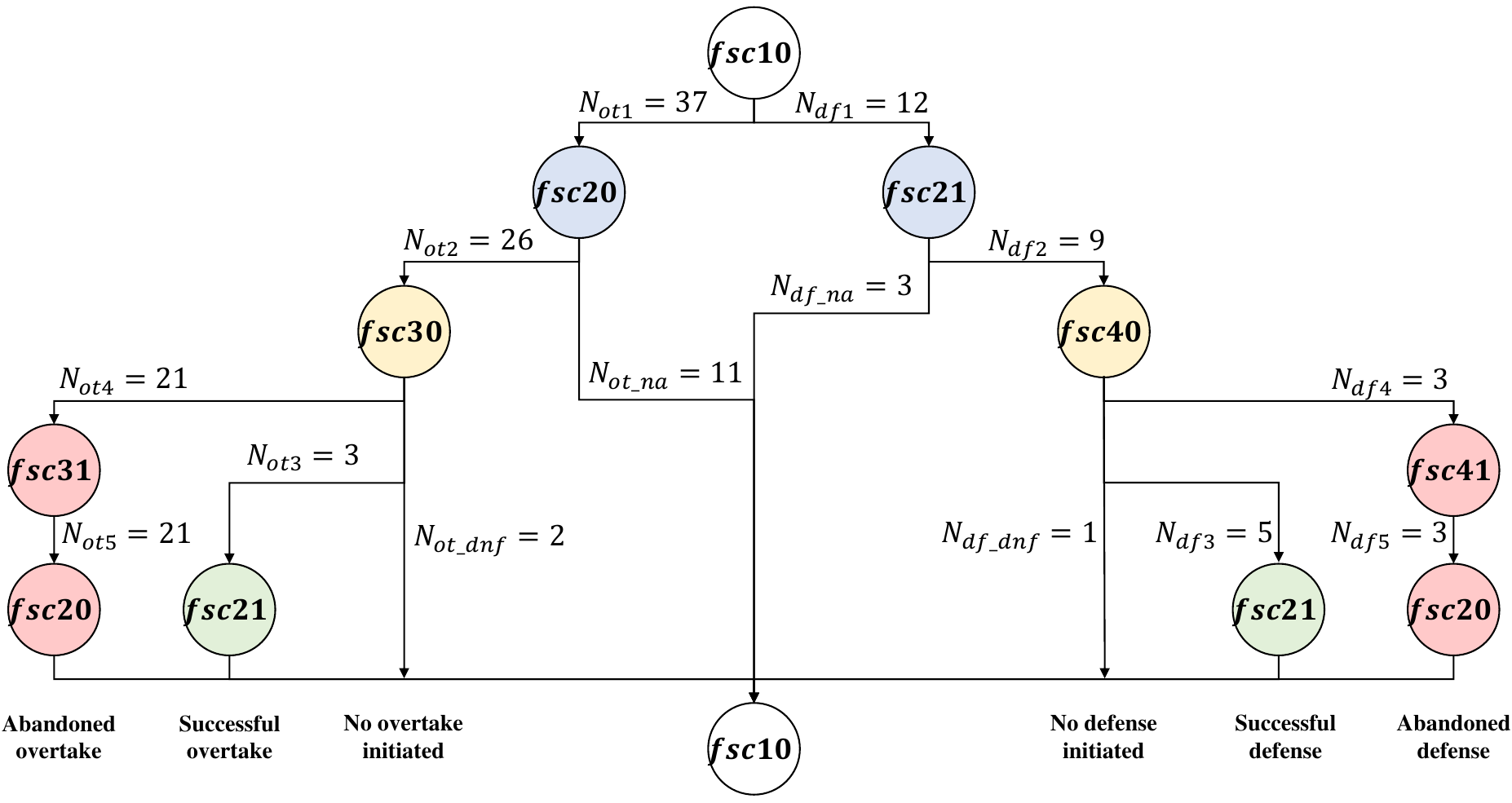}
    \caption{A numeric trace diagram showing the number of attempted, failed, incomplete and successful overtake and position defense attempts, and their associated traces. The individual traces sum up to the number of original attempts at an overtake or position defense, showing the completeness of the framework design.}
    \label{fig:trace_combo_result}
\end{figure*}

Figure ~\ref{fig:trace_combo_result} shows a numeric trace diagram for an experiment with the following parameters is shown in Figure~\ref{fig:trace_combo_result}: 25 lap head-to-head race, ego started in second position, and triggers $trig0$ = 150m, $trig1$ = 25m, $trig2$ = 30m, $trig3$ = 25m, $trig4$ = 20m, $trig5$ = 20m, $trig6$ = 6.0s, $trig7$ = 1.5s, $trig8$ = 7.5m.
From Figure~\ref{fig:trace_combo_result}: the ego initiated 26 overtake attempts: 3 successful, 21 abandoned, and 2 unfinished overtakes.
In the same experiment, the opponent was observed to have 9 position defense attempts: 5 successful, 3 failed, and 1 unfinished defense attempt.
Unfinished maneuvers resulted from the racecars leaving the passing zones.
The traces show that the number of successful overtake attempts made by one racecar is the same as the number of failed position defense attempts made by the other racecar in this experiment.

\begin{figure}[h!]
    \centering
    \includegraphics[width=\linewidth]{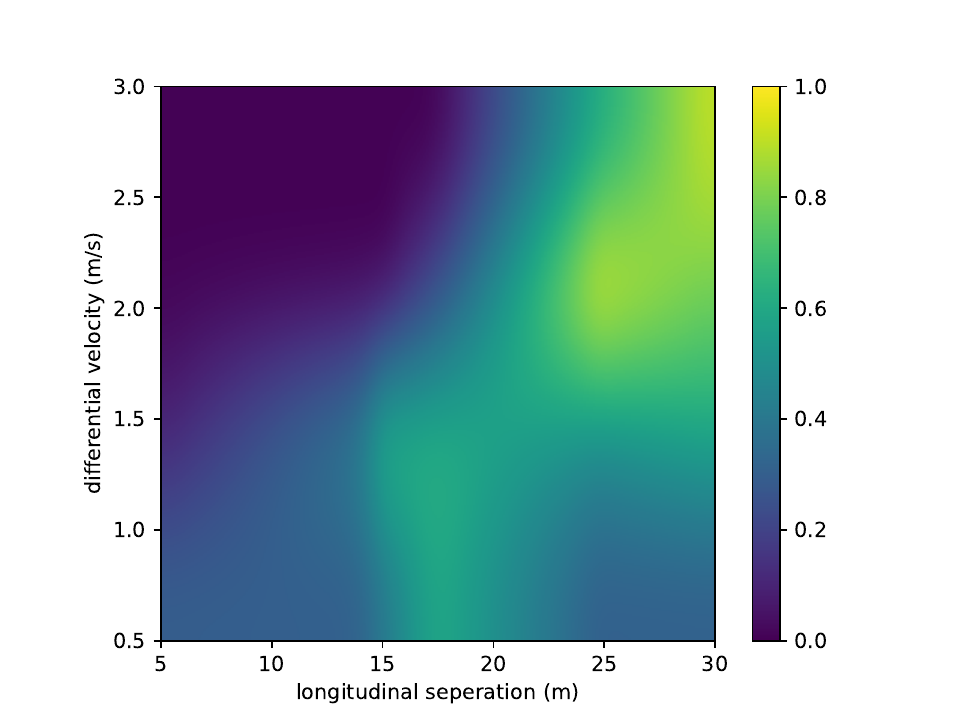}
    \caption{ARGOS overtake probability against a "mule" opponent that is not capable of defending its position.}
\label{fig:overtake_vs_mule}
\end{figure}

\begin{figure}[h!]
    \centering
    \includegraphics[width=\linewidth]{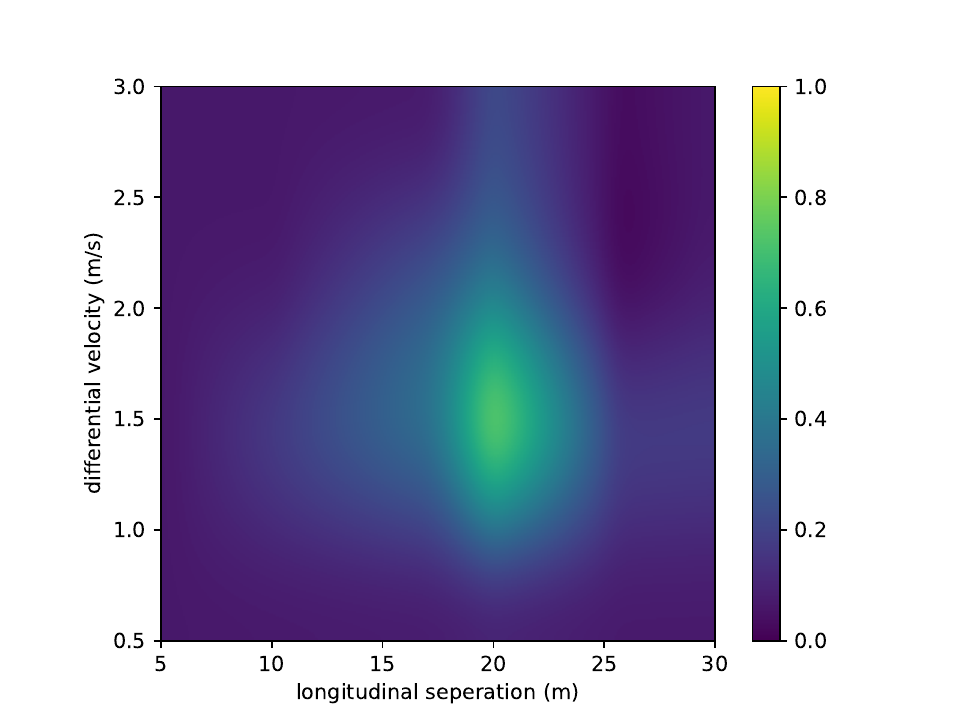}
    \caption{ARGOS overtake probability against an ARGOS equipped opponent that can block a pass.}
\label{fig:overtake_vs_ARGOS}
\end{figure}

\begin{figure}[h!]
    \centering
    \includegraphics[width=\linewidth]{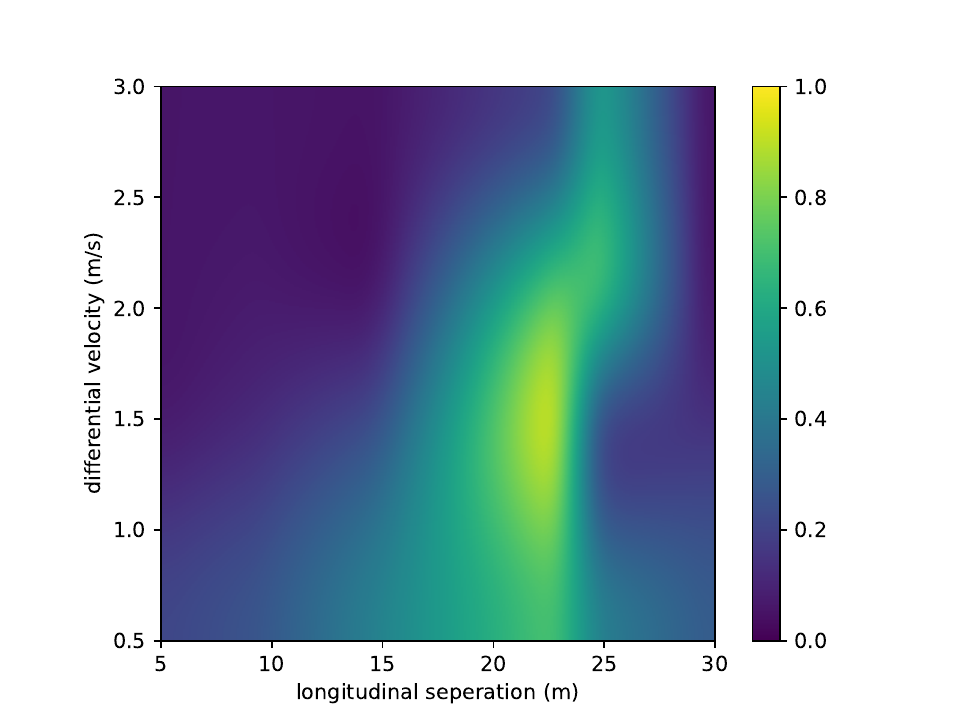}
    \caption{ARGOS position defense probability against an ARGOS equipped opponent that can overcome a block.}
\label{fig:defense_vs_ARGOS}
\end{figure}

Also from Figure~\ref{fig:trace_combo_result}, the working of the ARGOS framework as expected can be verified by observing the number of events in the trace sequence.
We observed that the number of initiations was equal to the sum of the number of successful and failed attempts for both overtake and position defense.
For example: in Figure~\ref{fig:trace_combo_result}, the left side of the trace diagram shows the overtake attempts of the ego racecar.
Here $N_{ot1}$ = 37 is the number of times the ego found itself in the passing zone behind the opponent, $N_{ot2}$ = 26 is the number of times the ego attempted to pass the opponent for one experiment session, $N_{ot3}$ = 3 is the number of successful overtake attempts and $N_{ot4, ot5}$ = 21 is the number of abandons and successful merge-backs due to a failed overtake attempt.
Finally, $N_{ot\_dnf}$ = 2 is the rare occurrence of race cars leaving the passing zone before a maneuver can be completed.
Equation~\ref{eq:design_verification} describes the summation property of the framework for both overtakes and position defense attempts for one experiment.

\begin{equation}
    \begin{split}
        N_{ot1} & = N_{ot2} + N_{o
t3} + N_{ot4, ot5} + N_{ot\_dnf} \\
        N_{df1} & = N_{df2} + N_{df3} + N_{df4, df5} + N_{df\_dnf}
    \end{split}
    \label{eq:design_verification}
\end{equation}

\section{Experiments}
\label{sec:experiments}
The experiments were conducted using the LGSVL simulator with the Indianapolis Motor Speedway (IMS) race-track and two similar Dallara IL-15 racecars modified to mimic the dynamics and performance of the AV-21 autonomous racecar used in the Indy Autonomous Challenge.

\noindent \textbf{Experiment Setup}: In this paper, an experiment is defined as a head-to-head autonomous racing session between two similar IL-15 racecars that lasted for 25 laps each and included one pit-out (drive from the pit area to the active race-track) and one pit-in (drive from the active race-track to the pit area) maneuver.
We dynamically adjusted the triggers listed in Table~\ref{tab:argos_triggers} to optimize overtakes and position defense, and observed the following.

\noindent \textbf{Overtake against mule opponent:} For an overtake against an opponent incapable of defending its position (i.e., the mule opponent), we found that the ego racecar attempting to pass must have a positive differential velocity of around 2.0m/s, and the ego racecar must initiate the overtake at about 25m behind the opponent.
If the ego's states are not similar to these values, we noticed that the overtake attempt led to instability (high positive differential velocity), or an incomplete overtake (large initiation distance behind the opponent).
The probability map for this situation is shown in Figure~\ref{fig:overtake_vs_mule}, and is very similar to the observations made in~\cite{autopass_gen_1}.

\noindent \textbf{Overtaking an ARGOS opponent:} When an opponent is capable of defending its position from an overtake attempt, we found that the relative differential velocity from the previous result always led to an abandoned overtake.
This is because the ego racecar came too close to the opponent and did not have enough distance to evade the block before the end of the passing zone.
In this situation, we discovered that a lower positive differential velocity of around 1.5m/s with an overtake initiation distance of 20m led to a higher chance of evading a block - leading to a higher probability of a successful overtake.
The probability map for this situation is shown in Figure~\ref{fig:overtake_vs_ARGOS}.

\noindent \textbf{Defending against an overtake:} When defending against an overtake attempt; we discovered that the a negative differential velocity of around 2.5m/s and a trigger distance of about 20m lead to the highest probability of a successful position defense maneuver.
The probability map for this situation is shown in Figure~\ref{fig:overtake_vs_ARGOS}.
The figure shows that it is easier to defend the race position than it is to overtake an opponent (comparing the high probability yellow regions of Figures~\ref{fig:overtake_vs_ARGOS} and~\ref{fig:defense_vs_ARGOS}).

While formal verification of the ARGOS framework (incl. model checking) helped with meeting the requirements specified for the framework, we observed for any non-compliance using telemetry from the two racecars.
Figure~\ref{fig:geometric_threshold_result} shows an overview of the relative positions of the racecar throughout the experiment conducted using the same parameters from which the trace diagram in Figure~\ref{fig:trace_combo_result} was derived.
From Figure~\ref{fig:geometric_threshold_result}, we see that most of the attempted overtakes and position defenses were executed in the optimal zone defined by $[trig3, trig4]$ without violating the safety-distance required by \textbf{Rule R3}.
Some maneuvers, however, were triggered or completed in the sub-optimal (gray) area, and this was found to be an artifact of the delays caused by the time-consuming path-planner, which will be replaced in the future iteration of the ARGOS framework.

\section{Conclusion}
\label{sec:conclusion}
In this paper, we presented the ARGOS framework - a modular autonomous head-to-head racing framework with specific guidelines on integrating/adapting components within the framework with a set of well defined requirements.
In addition, we also presented our solution to overtaking and position defense problems using a network of automatons that decompose the complex maneuvering involved in the respective maneuvers.
We formally verified the functioning of the ARGOS framework by model checking the framework against the defined requirements and using any counter-examples generated in the process to refine the framework.
Using parameter optimization, we found the regions of trigger values that led to the maximum number of successful overtakes and position defenses.

\printbibliography

\end{document}